\documentclass[conference]{IEEEtran}
\IEEEoverridecommandlockouts
\usepackage{cite}
\usepackage{amsmath,amssymb,amsfonts}
\usepackage{algorithmic}
\usepackage{graphicx}
\usepackage{textcomp}
\usepackage{xcolor}
\usepackage{hyperref}
\usepackage{tabularx}
\usepackage{booktabs}
\usepackage{multirow}
\usepackage{placeins}
\usepackage{subfigure}

\graphicspath{{img/}}

\def\BibTeX{{\rm B\kern-.05em{\sc i\kern-.025em b}\kern-.08em
    T\kern-.1667em\lower.7ex\hbox{E}\kern-.125emX}}
\begin{document}

\title{SmaAT-QMix-UNet: A Parameter-Efficient Vector-Quantized UNet for Precipitation Nowcasting
}

 \author{\IEEEauthorblockN{Nikolas Stavrou, Siamak Mehrkanoon\IEEEauthorrefmark{1}\thanks{*Corresponding author.} }
 \IEEEauthorblockA{\textit{Department of Information and Computing Sciences, Utrecht University, Utrecht, Netherlands} \\
 n.stavrou@students.uu.nl, s.mehrkanoon@uu.nl}
 }

\maketitle

\begin{abstract}
Weather forecasting supports critical socioeconomic activities and complements environmental protection, yet operational Numerical Weather Prediction (NWP) systems remain computationally intensive, thus being inefficient for certain applications. Meanwhile, recent advances in deep data-driven models have demonstrated promising results in nowcasting tasks. This paper presents SmaAT-QMix-UNet, an enhanced variant of SmaAT-UNet that introduces two key innovations: a vector quantization (VQ) bottleneck at the encoder–decoder bridge, and mixed kernel depth-wise convolutions (MixConv) replacing selected encoder and decoder blocks. These enhancements both reduce the model’s size and improve its nowcasting performance. We train and evaluate SmaAT-QMix-UNet on a Dutch radar precipitation dataset (2016–2019), predicting precipitation 30 minutes ahead. Three configurations are benchmarked: using only VQ, only MixConv, and the full SmaAT-QMix-UNet. Grad-CAM saliency maps highlight the regions influencing each nowcast, while a UMAP embedding of the codewords illustrates how the VQ layer clusters encoder outputs. The source code for SmaAT-QMix-UNet is publicly available on GitHub \footnote{\href{https://github.com/nstavr04/MasterThesisSnellius}{https://github.com/nstavr04/MasterThesisSnellius}}.
\end{abstract}


\section{Introduction}
Numerical Weather Prediction (NWP) models are computationally expensive, requiring large-scale simulations that are impractical for edge deployment or rapid ensemble forecasting. As a result, many researchers have turned to data-driven approaches, with deep learning models increasingly explored and adopted for weather forecasting over the past decade \cite{mehrkanoon2019deep, abdellaoui2021symbolic, wang2018predrnn++, trebing2020wind, stanczyk2021deep, aykas2021multistream}.
Early data-driven nowcasting mainly relied on recurrent sequence models such as RNNs, LSTMs, and GRUs to model temporal dynamics, but these approaches largely neglected the spatial structure of radar and satellite imagery. ConvLSTM mitigates this by embedding convolutions within LSTM gates, enabling joint learning of motion and morphology and outperforming optical-flow baselines for short-range forecasts \cite{shi2015convolutional}, while the PredRNN family extends this with spatio-temporal memory to better preserve storm structure at longer lead times \cite{wang2018predrnn++}. More recent approaches adopt attention-based architectures. SmaAT-UNet integrates attention and depthwise-separable convolutions within a U-Net backbone to capture multi-scale features efficiently \cite{trebing2021smaat}. It achieves competitive nowcasting performance while only using a quarter of the parameters of a standard U-Net variant. Other related models such as WF-UNet and STC-ViT further demonstrating the effectiveness of lightweight attention and transformer-style mechanisms for nowcasting \cite{kaparakis2023wf, saleem2024stc}.

Interpretability is essential for deploying deep models. Explainable-AI methods like Grad-CAM \cite{selvaraju2017grad} and UMAP \cite{mcinnes2018umap} provide human-readable insights into both individual predictions and overall model behavior. Recent XAI surveys in meteorology argue that combining such local and global views is essential for building practitioner trust and for debugging models before operational rollout \cite{mamalakis2020explainable}.


This paper introduces SmaAT-QMix-UNet, a compact evolution of SmaAT-UNet that achieves marginally improved accuracy and precision while reducing trainable parameters by 37.5\%. The improvement comes from two modifications: a discrete vector-quantization (VQ) bottleneck and MixConv blocks. A VQ module is inserted at the encoder–decoder bridge, replacing the latent feature map with nearest codeword indices to produce a compressed, noise-robust representation that also supports cluster-level interpretation. In addition, selected depthwise-separable convolutions in the encoder and decoder are replaced with MixConv layers that blend multiple receptive field sizes within a single block \cite{tan2019mixconv}, preserving multi-scale sensitivity while reducing redundancy. Interpretability is provided through Grad-CAM saliency maps and UMAP projections of the learned VQ codewords. Together, these changes improve efficiency, accuracy, and interpretability while remaining suitable for edge deployment.

\section{Related Work}

Several studies have built upon the classic UNet architecture \cite{ronneberger2015u} to tackle precipitation nowcasting. For instance, AA-TransUNet \cite{yang2022aa} uses a transformer and a UNet with attention modules and depthwise-separable convolutions, demonstrating improved performance on precipitation nowcasting. Similarly, Broad-UNet \cite{fernandez2021broad} refines the UNet backbone with multi-scale feature extraction via asymmetric parallel convolutions and Atrous Spatial Pyramid Pooling (ASPP) module, yielding more accurate predictions with fewer parameters. Variants extending the SmaAt-UNet \cite{trebing2021smaat} approach include GA-SmaAt-GNet \cite{reulen2024ga}, which leverages a generative adversarial framework and integrates precipitation masks to boost performance in extreme events, and SAR-UNet \cite{renault2023sar}, which introduces residual connections alongside depthwise-separable convolutions to enhance both accuracy and interpretability through visual explanations. Lastly, WF-UNet \cite{kaparakis2023wf} took a different approach by fusing additional meteorological inputs via a 3D-UNet architecture. Other deep learning approaches have also been proposed for precipitation nowcasting beyond the UNet family. Early work using ConvLSTM has evolved into models like TrajGRU \cite{shi2017deep}, which learn location-variant recurrent connections to better capture natural motion. Moreover, models like MetNet \cite{sonderby2020metnet} and Nowcasting-Nets \cite{ehsani2021nowcasting} combine self-attention and recurrent structures to extend forecast horizons and deliver probabilistic predictions that capture uncertainty. The authors in \cite{vatamany2025graph} formulated precipitation nowcasting as a spatiotemporal graph sequence problem.

Various studies have demonstrated the effectiveness of VQ techniques across a range of applications, highlighting their potential to enhance model robustness, efficiency, and representational power. In the medical imaging domain, work has been done which shows that integrating a quantization block into a UNet architecture leads to improved segmentation accuracy and robustness against noise, domain shifts, and other perturbations by learning a discrete, low-dimensional representation \cite{santhirasekaram2022vector}. Similarly, VQ-UNet \cite{he2024vqunet} applies a multi-scale hierarchical VQ approach to defend deep neural networks against adversarial attacks, effectively reducing unwanted noise and reconstructing data with high fidelity. The widely known VQ-VAE paper, further establishes that replacing continuous latent codes with a learned discrete codebook can avoid issues like posterior collapse, ultimately enabling high-quality generation in images, videos, and speech \cite{yan2021videogpt}. Finally, the aforementioned GPTCast model \cite{franch2024gptcast}, which adapts the VQ-GAN framework \cite{esser2021taming}, illustrates that vector quantization can be effectively employed in precipitation nowcasting, in their case with a variational autoencoder, to generate accurate, high-resolution forecasts.

Beyond the depthwise-separable convolutions employed in SmaAT-UNet, efficient convolutions continue to diversify. MixConv partitions channels and applies several kernel sizes within one depth-wise layer, improving the accuracy-to-FLOPs ratio on mobile-scale models \cite{tan2019mixconv}, while GhostConv (from GhostNet) generates “ghost” feature maps through inexpensive linear operations, slashing parameter count without sacrificing representational power \cite{han2020ghostnet}. 



\section{Method}

\begin{figure*}[h!]
    \centering
    \includegraphics[width=\linewidth]{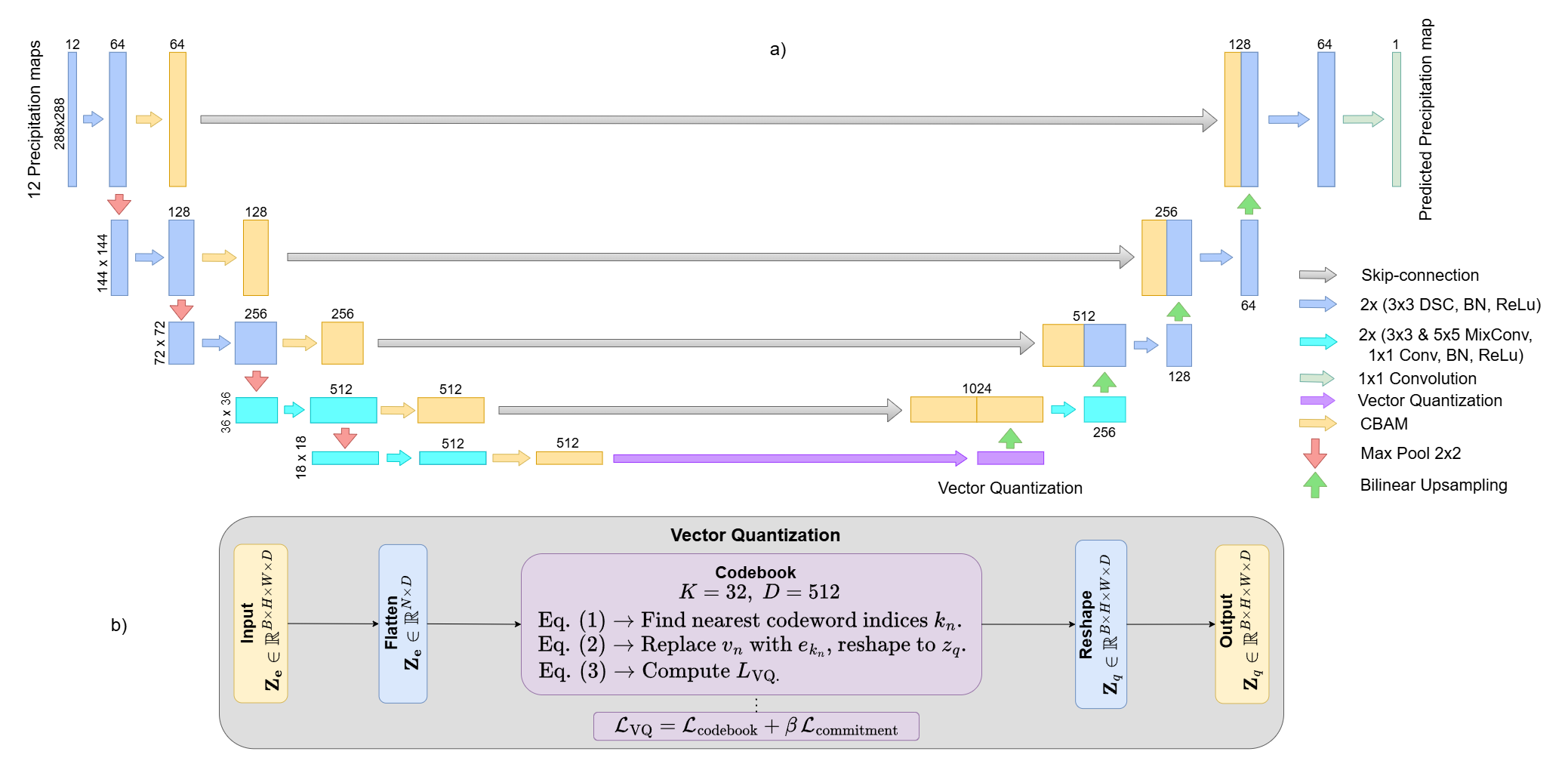}
    \caption{
    (a) \textbf{SmaAT-QMix-UNet architecture:} Rectangles represent feature maps, with height indicating spatial resolution and width the channel dimension. MixConv blocks are used in the last two encoder levels and the first decoder stage, while a VQ layer discretizes the $B \times 18 \times 18 \times 512$ bottleneck tensor.  
(b) \textbf{Vector-quantization module:} Latent features are flattened, each 512-D vector is assigned to its nearest codebook entry ($K=32$), and reshaped into a quantized feature map. Training optimizes the combined codebook and $\beta$-weighted commitment losses.}
    \label{fig:fullwidth}
\end{figure*}

\subsection{Proposed SmaAT-QMix-UNet model}

\subsubsection{Architecture}

Fig. \ref{fig:fullwidth} sketches the proposed SmaAT-QMix-UNet, which follows the encoder–bottleneck–decoder template of SmaAT-UNet \cite{trebing2021smaat} but introduces two key modifications. The encoder comprises five hierarchical levels (blue and cyan arrows). In each level, two 3 × 3 depth-wise separable convolutions, BatchNorm and ReLU are followed by a CBAM attention module. The CBAM output is forwarded via a skip connection (grey arrows) to the matching decoder stage, while a 2×2 max pool (red arrow) halves the spatial resolution and feeds the next level. Levels 1–3 use the original depthwise-separable convolutions while Levels 4 and 5 use a MixConv block (cyan arrows) that processes channel groups with 3×3 and 5×5 kernels in parallel \cite{tan2019mixconv}. At the bottleneck between encoder and decoder, the last encoder tensor is routed through a VQ module (purple arrow). Each latent vector is replaced by its nearest codeword entry in a learned codebook and then passed to the decoder. The decoder mirrors the encoder with four stages. Each begins with bilinear up-sampling (green arrows) that doubles spatial dimensions and concatenates the result with the corresponding skip connection. The first decoder stage reuses the MixConv block, while the remaining stages revert to depth-wise separable convolutions. A final 1×1 convolution (purple arrow) produces the single-channel precipitation nowcast at 30 mins lead time. 

\subsubsection{Vector-Quantization (VQ) Module}
On the bottleneck of our model architecture, we use a VQ module following the lines of VQ-VAE, \cite{yan2021videogpt}.
The continuous encoder output
$\mathbf{z}_e\!\in\!\mathbb{R}^{B\times H\times W\times C}$
is mapped to a discrete latent space defined by a learnable
\emph{codebook}
$\mathcal{E}=\{\mathbf{e}_k\}_{k=1}^{K}$,
with codeword $\mathbf{e}_k\!\in\!\mathbb{R}^{D}$. 
We choose the codeword dimensionality $D$ to match the number of channels $C$, i.e., $D=C$. Therefore, in the rest of the paper, we use $D$ to also indicate the number of channels. As in \cite{van2017neural}, the codebook $\mathcal{E}$ is a set of learnable parameters, optimized jointly with the rest of the model. Next we flatten $\mathbf{z}_e$ to a matrix $\mathbf{z}_e \in \mathbb{R}^{N\times D}$, where $N = B \times H \times W$. We denote every row of this matrix as $\mathbf{v}_n \in \mathbb{R}^{D}$. For each vector $\mathbf{v}_n$, we compute the squared $\ell_2$ distance to every codeword and select the index of the nearest codeword as follows:
\begin{equation}
k_n=\underset{k \in \{1,\dots,K\}}{\arg\min}
     \bigl\lVert \mathbf{v}_n - \mathbf{e}_k \bigr\rVert_2^{2},
\qquad \textrm{for}\;\; n = 1,\dots,N.
\label{eq:nearest_code}
\end{equation}
Next, each $\mathbf{v}_n$ is replaced with its corresponding codeword $\mathbf{e}_{k_n}$, and the result is reshaped to match the original layout, producing the quantized tensor as follows:
\begin{equation}
\mathbf{z}_q=\textrm{reshape}
        \bigl(\{\mathbf{e}_{k_n}\}_{n=1}^{N}\bigr),
        \quad
\mathbf{z}_q \in \mathbb{R}^{B\times H\times W\times D}.
\label{eq:quantised}
\end{equation}
Incorporating the VQ module introduces two additional loss terms, commitment loss and codebook loss which are computed during training. Commitment loss penalizes encoder vectors for drifting away from their selected code embeddings and thus encourages them to “commit” stably to a discrete code and codebook loss, which pulls each codeword in $\mathcal{E}$  toward its detached encoder output to keep the dictionary representative. Training uses the straight-through estimator together with the two-term loss:
\begin{equation}
\mathcal{L}_{\mathrm{VQ}}
= \frac{1}{N}  \sum_{n=1}^{N}
    \underbrace{\bigl\lVert \mathrm{sg}[\mathbf{v}_n] - \mathbf{e}_{k_n}\bigr\rVert_2^2}_{\text{codebook}}
  \;+\;
   \frac{\beta}{N}\sum_{n=1}^{N}
    \underbrace{\bigl\lVert \mathbf{v}_n - \mathrm{sg}[\mathbf{e}_{k_n}]\bigr\rVert_2^2}_{\text{commitment}} ,
\label{eq:vq_loss_detailed}
\end{equation}

where $\operatorname{sg}[\cdot]$ is the stop-gradient operator and
$\beta$ is the commitment cost controlling how strongly the encoder is
encouraged to commit to its selected codes. All terms are computed as the mean squared error over all latent elements, making the loss magnitude insensitive to batch size or spatial resolution. At inference time the module is deterministic, mapping each encoder activation to the nearest codeword entry in $\mathcal{E}$.

\subsubsection{Mixed Convolution Block}



\begin{figure}[h!]
    \centering
    \includegraphics[width=0.7\linewidth]{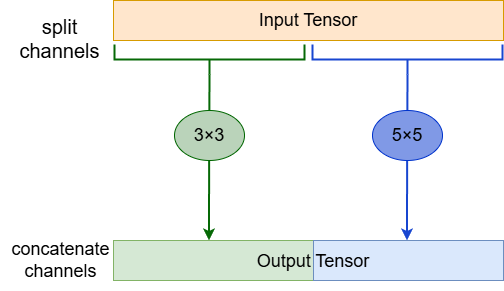}
    \caption{Mixed depthwise convolution (MixConv). The input tensor is split in two disjoint groups. First group is processed by a 3x3 depthwise convolution and the second group by a 5x5 depthwise convolution. The two outputs are then concatenated along the channel dimension.}
    \label{fig:mixconv}
\end{figure}

In the original SmaAT-UNet, each DoubleDSC unit stacks two depth-wise separable convolutions with fixed 3×3 kernels. Since the deepest encoder layers have the widest channel dimensions and require the largest receptive fields, we modify only these layers and the first decoder stage after the VQ bottleneck. Specifically, we replace the DoubleDSC units in the two deepest encoder levels and the first decoder stage with a double MixConv block. Each MixConv splits the feature map into two equal channel groups, applies 3×3 and 5×5 depth-wise convolutions, concatenates the results, and projects them through a shared 1×1 point-wise convolution with BN and ReLU (Fig.~\ref{fig:mixconv}). This sequence is repeated twice to mirror the original DoubleDSC structure. The resulting design captures both local and broader spatial context while reducing parameter count, and is applied only at these three locations to preserve the baseline behaviour elsewhere.

\subsubsection{Model Variation}
To showcase the effects of VQ and MixConv we evaluate four progressively modified networks. The starting point is the unaltered SmaAT-UNet, which acts as the reference baseline. Adding MixConv alone, restricted to the last two encoder levels and the first decoder stage, yields SmaAT-Mix-UNet, a purely convolutional variant that probes the impact of kernel diversity on accuracy and size. Additionally, we train SmaAT-Q-UNet, which inserts the VQ bottleneck but leaves all depth-wise separable convolution layers unchanged, isolating the contribution of discretized latents. Finally, we combine both modifications in SmaAT-QMix-UNet. This model includes the VQ layer and the same three MixConv replacements used in SmaAT-Mix-UNet, providing the configuration that jointly targets compactness and accuracy. All four networks share identical training and optimizer settings, and early-stopping criteria, ensuring that any performance differences can be attributed to the architectural changes alone.

\subsection{Training}

For our training, we follow the same steps as \cite{trebing2021smaat}. Our model ingests twelve consecutive radar maps per step and is trained for at most 100 epochs on a single NVIDIA H100 GPU with the Adam optimiser and an initial learning rate of 0.001 and a learning rate patience of 4 which reduces learning rate if no improvement on validation loss is shown for 4 consecutive epochs. If the validation loss shows no improvement for 15 straight epochs, training stops early. Through hyperparameter tuning, we set a codebook length of 32 and a commitment cost of 0.75 that give the best model performance.

\subsection{Evaluation}
Primary performance is reported as mean-squared error (MSE) over the test split. To assess event detection quality, each output is thresholded into rain/no-rain masks, counts of true positives, false positives, true negatives and false negatives then yield precision, recall, accuracy and F1-score. Results for every SmaAT-QMix-UNet variant are benchmarked against the unaltered SmaAT-UNet and a Persistence baseline that simply repeats the last input frame.

\subsection{Explainability}
We pair a local and a global tool to scrutinise model behavior. Gradient-weighted Class Activation Mapping (Grad-CAM) is applied to every encoder and decoder level to highlight the pixels that most influence each nowcast horizon. Complementing these saliency maps, Uniform Manifold Approximation and Projection (UMAP) embeds the full set of vector-quantized code indices into two dimensions, revealing how discrete codes cluster into recurring weather regimes and allowing their associated Grad-CAM maps to be inspected side by side. Together, Grad-CAM and UMAP provide a coherent view of how the network combines spatial cues and latent code patterns to produce its predictions.

\begin{table*}[htbp]
\centering
\caption{
Performance comparison at 30-min lead time on the NL-50 test set, including persistence, SmaAT-UNet, and proposed models, with model size and inference time reported. Best values are in bold
}
\begin{tabularx}{\textwidth}{%
    >{\raggedright\arraybackslash}X
    >{\centering\arraybackslash}p{1.5cm}
    >{\centering\arraybackslash}p{1.5cm}
    >{\centering\arraybackslash}p{1.5cm}
    >{\centering\arraybackslash}p{1.5cm}
    >{\centering\arraybackslash}p{1.5cm}
    >{\centering\arraybackslash}p{1.5cm}
    >{\centering\arraybackslash}p{1.5cm}
    }
\toprule
\textbf{Model}
  & \textbf{Parameters}
  & \textbf{Inference Time (ms)}
  & \textbf{MSE (px)}
  & \textbf{Precision}
  & \textbf{Recall}
  & \textbf{Accuracy}
  & \textbf{F1 Score} \\
\midrule
Persistence       & –             & –      & 0.0248          & 0.678           & 0.643          & 0.756           & 0.660         \\
SmaAT-UNet        & 4 M           & 45   & 0.0122          & 0.730           & 0.850          & 0.829           & 0.786         \\
SmaAT-Q-UNet      & 4 M           & 41   & \textbf{0.0119} & 0.748           & 0.820          & 0.832           & 0.782         \\
SmaAT-Mix-UNet    & \textbf{2.5 M} & 42   & 0.0129          & 0.670           & \textbf{0.866} & 0.794           & 0.756         \\
SmaAT-QMix-UNet   & \textbf{2.5 M} & \textbf{39} & 0.0120   & \textbf{0.763}  & 0.812          & \textbf{0.838}  & \textbf{0.787}\\
\bottomrule
\end{tabularx}
\label{tab:6step_results_nl50}
\end{table*}

\section{Experiments}

We use the KNMI precipitation dataset from \cite{trebing2021smaat}, comprising approximately 420 000 radar composites recorded every five minutes between 2016 and 2019 by the Dutch C-band radars at De Bilt and Den Helder. Images are normalised, cropped to the common radar coverage, and centre-cropped to 288 × 288 pixels. Following \cite{trebing2021smaat}, only samples with at least 50\% rainy pixels in the target frame are retained, forming the NL-50 subset used for all experiments. Each sample consists of twelve consecutive rain maps (60 min history), with the model predicting precipitation 30 min ahead using mean squared error as the loss. Training follows the baseline setup with Adam, batch size 8, and identical learning-rate scheduling and early stopping. The only additional hyperparameters are the VQ codebook size $K$ and commitment cost $\beta$, tuned on the validation set; the best configuration uses $K=32$ and $\beta=0.75$. The model with the lowest validation loss is evaluated on the NL-50 test set.

\begin{figure}[t]
  \centering

  \includegraphics[width=0.9\linewidth,clip,trim=0 0 567 0]{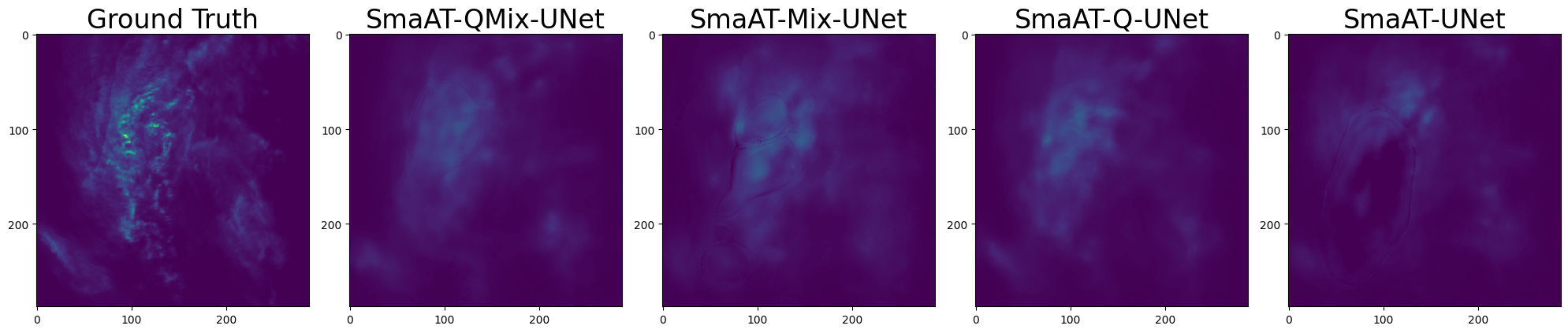}\\[-1mm]

  \includegraphics[width=0.57\linewidth,clip,trim=880 0 0 0]{img/image_predictions}

  \caption{Comparison of predictions generated by different models. The SmaAT-QMix-UNet model shows better alignment with the ground truth.}
  \label{fig:predictions}
\end{figure}

\begin{figure*}[t!]
  \centering
  \subfigure[]{
    \includegraphics[width=0.64\textwidth]{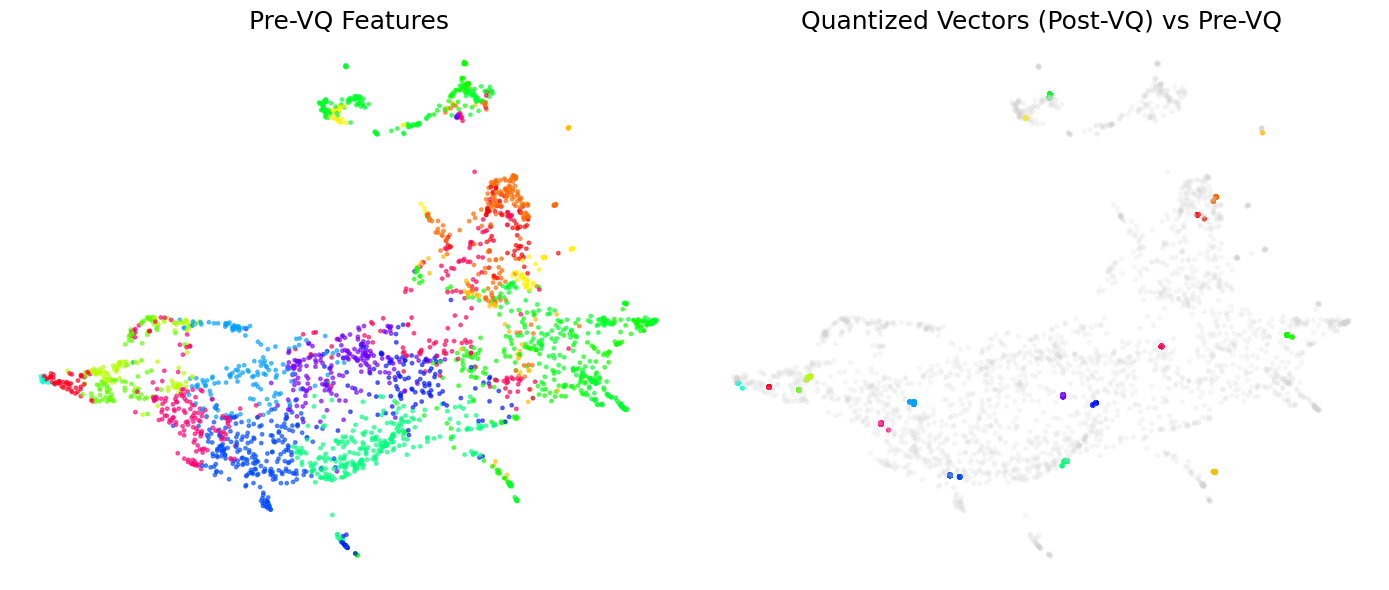}
    \label{fig:umap}
  }\hfill
  \subfigure[]{
    \includegraphics[width=0.24\textwidth]{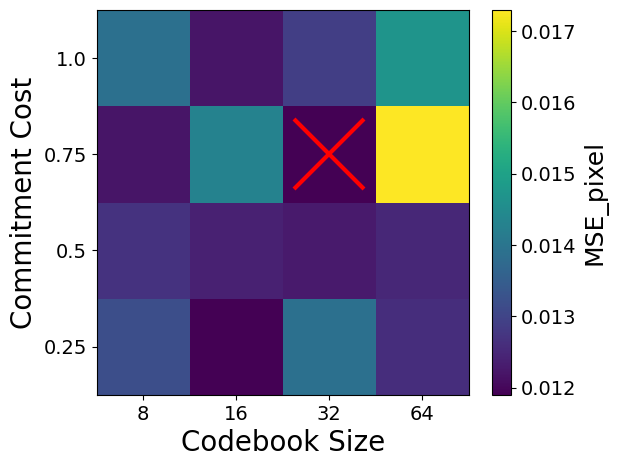}
    \label{fig:hyperparametertuning}
  }
  \caption{
  (a) UMAP visualization of encoder feature vectors before and after vector quantization in SmaAT-QMix-UNet, where grey points denote pre-VQ representations and colored points indicate their assigned codewords. (b) Hyperparameter tuning results for the VQ module, showing validation performance across 16 combinations of codebook size and commitment cost, with $K=32$ and $\beta=0.75$ achieving the best performance.}
\end{figure*}

\begin{figure*}[t!]
    \centering
    \includegraphics[width=0.8\linewidth]{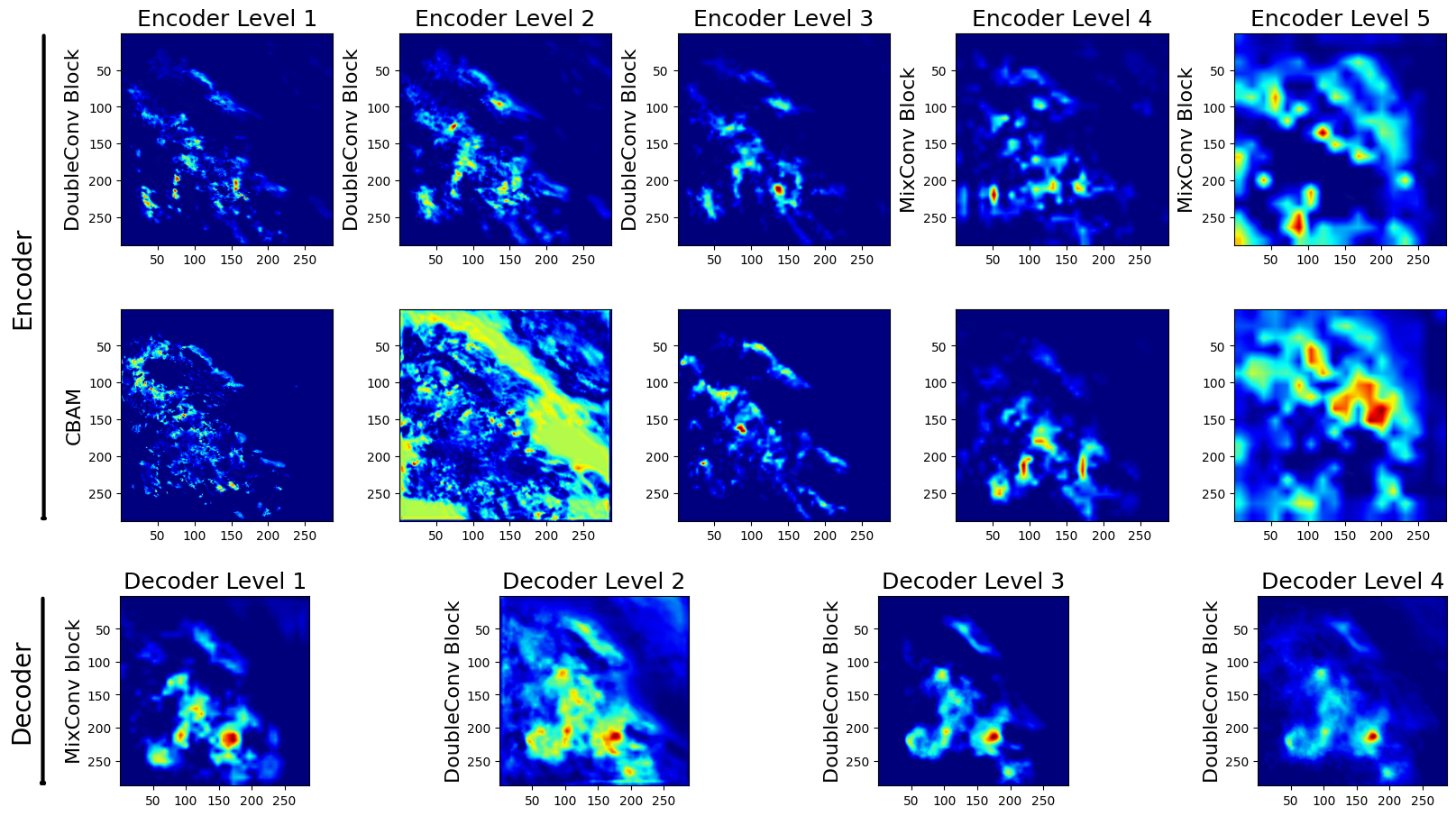}
    \caption{
    Heatmaps generated with Grad-CAM for SmaAT-QMix-UNet, showing activation regions across the five encoder and four decoder levels, including responses from the convolutional blocks (DoubleDSC or MixConv) and the CBAM modules in the encoder.
    }
    \label{fig:gradcam_encoder}
\end{figure*}

\section{Results and Discussion}

\subsection{Precipitation nowcasting}

\subsubsection{Tuning}

We tune the two VQ-specific hyperparameters, the codebook size $K$ and the commitment cost $\beta$, using a coarse grid search with $K \in \{8,16,32,64\}$ and $\beta \in \{0.25,0.50,0.75,1.00\}$ on the NL-50 validation set. Fig.~\ref{fig:hyperparametertuning} shows the resulting pixel-wise MSE heatmap. The best performance is obtained at $K=32$ and $\beta=0.75$, with $K=16$, $\beta=0.25$ yielding comparable results. Overall, moderate codebook capacity combined with a relatively high commitment cost provides the best balance between representation diversity and quantization stability. We therefore fix $K=32$ and $\beta=0.75$ for all subsequent SMiQ-UNet experiments.

\vspace{0.5\baselineskip}
\subsubsection{Evaluation}

The experimental results on the Dutch precipitation dataset demonstrate the advantages of SmaAT-QMix-UNet in both predictive performance and model compactness. Table~\ref{tab:6step_results_nl50} reports 30-minute lead-time results on the NL-50 test set, including accuracy metrics, model size, and runtime. In terms of MSE, SmaAT-Q-UNet and SmaAT-QMix-UNet slightly outperform the SmaAT-UNet baseline (0.0119 and 0.0120 vs.\ 0.0122), whereas SmaAT-Mix-UNet alone underperforms (0.0129), indicating that MixConv without discretization is insufficient. All learned models outperform the persistence baseline by a large margin. Across secondary metrics, SmaAT-QMix-UNet matches or exceeds the baseline on all scores except recall (0.812 vs.\ 0.850), likely due to VQ regularization suppressing weak precipitation cells. This is offset by higher precision (+0.033) and improved overall accuracy. Crucially, SmaAT-QMix-UNet achieves these results with only 2.5M parameters, 37.5\% fewer than the baseline, and reduces inference time by approximately 6\,ms per batch. Overall, the results confirm that selective MixConv drives parameter efficiency, while the VQ bottleneck preserves or improves skill, yielding a compact and efficient nowcasting model. Figure~\ref{fig:predictions} provides a qualitative comparison of 30-minute forecasts, showing that SmaAT-QMix-UNet produces predictions closest to the ground truth, particularly in regions of higher precipitation intensity.
\subsection{UMAP visualization}
Fig.~\ref{fig:umap} uses Uniform Manifold Approximation and Projection (UMAP) to visualize the effect of vector quantization on the encoder latent space. The left panel shows a two-dimensional embedding of the 512-D encoder features, colored by their assigned codeword. After quantization (right), features collapse onto a small set of discrete codewords, with colored points indicating codeword locations and grey points showing the original feature positions. The tight clustering demonstrates that the codebook efficiently compresses similar patterns while preserving the overall latent-space structure.

\subsection{Grad-CAM visualization}

Fig.~\ref{fig:gradcam_encoder} presents Grad-CAM saliency maps for all encoder and decoder levels of SmaAT-QMix-UNet, showing heatmaps for the convolutional blocks and the corresponding CBAM units. In the shallow encoder (Levels 1–3), DoubleDSC blocks already outline the main precipitation regions, while CBAM distributes attention more broadly. By Level 3, both modules converge on the central rain areas. In deeper encoder layers (Levels 4–5), MixConv maintains focus on precipitation structures, with CBAM highlighting complementary regions as representations become more abstract. During decoding, saliency initially remains concentrated on high-intensity precipitation, then expands through up-sampling and skip connections before progressively refocusing on the heaviest rainfall. Overall, the saliency maps indicate a hierarchical representation in which SmaAT-QMix-UNet captures global rainfall geometry in early layers, emphasizes intense precipitation in deeper layers, and refines this information during decoding, while MixConv preserves spatial localization despite larger receptive fields.

\section{Conclusion}

In this work, we introduced SmaAT-QMix-UNet, which combines a vector-quantization bottleneck with MixConv to preserve the multi-scale design of SmaAT-UNet while substantially reducing model size. By discretizing the latent space and replacing the deepest convolutional blocks with MixConv, the model achieves marginal improvements in nowcasting skill with significantly fewer parameters, making it well suited for resource-constrained and edge deployments. Interpretability is enhanced through a two-level analysis: Grad-CAM highlights spatial regions driving the 30-minute forecast, while UMAP projections of the VQ codewords reveal coherent latent-space clusters. Together, these properties reduce inference and training costs and support efficient, interpretable precipitation nowcasting.

\FloatBarrier
\bibliographystyle{IEEEtran}
\bibliography{references-latest}

\vspace{12pt}

\end{document}